\documentclass[11pt,a4paper]{article}

\usepackage[T1]{fontenc}
\usepackage[utf8]{inputenc}
\usepackage{lmodern}
\usepackage[margin=2.4cm]{geometry}
\usepackage{graphicx}
\makeatletter
\@ifundefined{KV@Gin@alt}{\define@key{Gin}{alt}{}}{}
\makeatother
\usepackage{booktabs}
\usepackage{amsmath}
\usepackage{microtype}
\usepackage[section]{placeins}
\usepackage[table]{xcolor}
\usepackage{enumitem}
\usepackage[font=small,labelfont=bf,skip=6pt]{caption}
\usepackage{natbib}
\usepackage[hidelinks,breaklinks]{hyperref}

\graphicspath{{figures/pdf/}}
\setcitestyle{authoryear,round}

\definecolor{axink}{HTML}{1A1A1A}
\definecolor{axmuted}{HTML}{5C5953}
\definecolor{axrule}{HTML}{D6CFBC}

\makeatletter
\renewcommand\paragraph{\@startsection{paragraph}{4}{\z@}%
  {1.5ex \@plus .4ex \@minus .2ex}%
  {-0.9em}%
  {\normalfont\normalsize\bfseries}}
\makeatother

\newcommand{\n}[1]{\textbf{#1}}
\providecommand{\noopsort}[1]{}

\newsavebox{\tabbox}
\newcommand{\tabnote}[1]{%
  \\[4pt]\begin{minipage}{\wd\tabbox}\footnotesize\textcolor{axmuted}{#1}\end{minipage}}
\newcommand{\widetabnote}[1]{%
  \\[6pt]\begin{minipage}{\linewidth}\footnotesize\textcolor{axmuted}{#1}\end{minipage}}

\title{\textbf{AX is the New AEO}}

\author{%
  \begin{tabular}{c@{\hspace{3em}}c@{\hspace{3em}}c@{\hspace{3em}}c}
    Ido Finder & Assaf Elovic & Gad Shalev & Liad Yosef \\
    \small\texttt{ido@ora.ai} & \small\texttt{assaf@ora.ai} & \small\texttt{gad@ora.ai} &
    \small\texttt{liad@ora.ai}
  \end{tabular}%
}
\date{\normalsize ora research}

\begin{document}
\maketitle

\begin{abstract}
\noindent
In 2023, AI models answered from training data and hallucinated when it ran out, and businesses were
told to seed that knowledge. Models' training knowledge has since given way to live web search, and
the advice followed it there: answer-engine optimization, or AEO, now tells businesses to scatter
breadcrumbs across forum threads, listicles, and off-site citations, so AI engines are likelier to
surface and recommend them. But being surfaced is no longer enough: an agent opens the results and
reads them before deciding, and one buyer question sends it through several rounds of search and
fetch. What
decides the outcome at this drill-down step is whether the agent can fetch and read the business's
own site: \textbf{agent experience (AX)}. We argue that AX is the new AEO. We
run \n{37,927} agent journeys, each a buyer question about a business, across
\n{four independent harnesses} over \n{1,056 real businesses}, matched on fame, prior model
knowledge, and two AEO proxies, then split based on their AX level. Only \n{7--10\%} of the
finished answer comes from the model's training knowledge, whether or not the site is readable. Agent-ready businesses have answers built from their
own pages \n{78\%} of the time against \n{56\%} and are \n{clearly recommended 1.9$\times$} more
often, while a grounded answer about a not-agent-ready business costs the agent
\n{64\% more} on average. Holding business, harness, and question fixed, answers built from the site are
\n{41\% more accurate} on average. The dominant failure is not fabrication but \emph{omission}: web-built
answers are \n{3.7$\times$} more likely to contain none of the facts the buyer asked for. Baselines
differ sharply across the four harnesses, with clear-recommendation rates varying \n{sevenfold}
from stack to stack, yet the recommendation gap holds in every one. In the agentic web era, being readable
beats being talked about, and \n{improving a site's AX} is the strongest lever a business has.
\end{abstract}

\section{Introduction}

For two decades the web has been written for human readers and for the search crawlers that index
pages on their behalf: layouts that reward attention, interfaces that answer to clicks, content
assembled in the browser after the page loads. The arrangement held because the two audiences
wanted compatible things. A crawler had to find the page and gather roughly what it was about; a
person did the rest, in a browser, at human speed.

That is no longer who visits. As of September 2026, bots make \n{62.4\%} of requests for HTML
content on one major network, against \n{37.6\%} from people \citep{cloudflare2026radartraffic}.
Much of that is the search and SEO machinery that has crawled the web for years, but the part
growing fastest is new. AI answer engines like ChatGPT, Claude, and Gemini now reply to a question
in place of a page of links, and
agents search, fetch, and read on a person's behalf before reporting
back \citep{gou2025mind2web2,steiner2026interfaces}. These are not crawlers building an index for later.
They arrive because someone has just asked something, and they read in order to answer it.

They consume the page on different terms. Most do not execute JavaScript by default, so content
assembled after load is often invisible to them \citep{vercel2024crawler}; they rarely browse a
site so much as fetch a URL; and what they cannot parse into text tends to be discarded. A page
that is perfectly legible to a person can be close to empty to an agent. In early 2026, ora
introduced the agent-readiness ranker we use in this study, which scores a site against agent
protocols and observed agent behaviour \citep{ora2026score}. The ranker is our own, and we
report its scores as the study's instrument rather than as an independent measure. Across nearly \n{100,000}
sites, fewer than \n{1\%} earn its top grade of A or above: most of the web is not ready for the consumers it now
has, and a standards effort is under way to define what readiness requires
\citep{anthropic2024mcp,microsoft2025nlweb,agentready2026}.

\citet{biilmann2025ax} coined \textbf{agent experience (AX)} for the experience an agent has as
the user of a product or platform. We apply it to the site
itself: a site's AX is how well it serves the agent that arrives to fetch and read it, and a site
with good AX is what the ranker scores as agent-ready. Our claim is that AX is what decides
whether a business's own facts reach the buyer.

The optimization industry has followed the traffic. When models answered largely from what they had
learned in training, the advice was to get into the corpus itself, seeding the sources a model
would learn from. As models began retrieving at answer time rather than recalling, the
advice moved with them, and answer-engine optimization (AEO) and generative engine optimization
(GEO) formalized the retrieval-era version \citep{aggarwal2024geo}. The two labels are often used
interchangeably, and for simplicity we write AEO throughout for both. The current playbook
tells businesses to scatter breadcrumbs where a retrieval step will find them, in forum threads,
listicles, and off-site citations, so that an AI engine is likelier to surface and recommend them.
Classic search optimization has not gone away either: agents reach most pages through a web search,
so what ranks still decides much of what an agent sees, and a benchmark of conversational-SEO
rewrites finds most of them ineffective or actively harmful to a source's ranking, with classic SEO
more effective \citep{puerto2025cseo}.

But being surfaced is no longer the end of the process. A single buyer question sends an agent
through several rounds of search and fetch, and what it reads in those rounds is what the answer
gets built from. SEO and current AEO both govern the surfacing step; the drill-down step decides
what the answer is made of. If the agent can read you, it grounds its answer in your
first-party facts. If it cannot, because the page is JavaScript-only, because the site's bot controls block the
agent \citep{cloudflare2025default}, or because access is otherwise
gated \citep{longpre2024consent}, the agent does not stop. It has the open web to fall back on:
third-party pages, competitors, listicles, aggregators. Whether that substitution happens, how
often, and what it costs is what this paper measures.

AX is broader than reading. An open agent readiness specification separates three layers, whether an agent can
find a site, whether it can read it, and whether it can act on it through documented APIs or an MCP
server \citep{agentready2026}. We scope this study to the read layer, holding discovery fixed and
leaving the act layer to later work. Our thesis is that AX is the new AEO: answer-engine
optimization decomposes into SEO for the surfacing step and AX for the drill-down step and
everything after it, and at the drill-down being readable beats being talked about. The main
contributions are:

\begin{enumerate}[leftmargin=1.4em,itemsep=2pt,topsep=3pt]
  \item \textbf{The study runs the first controlled experiment on agent readiness in the wild} we
    know of: \n{37,927} agent journeys on \n{1,056} live businesses across four harnesses.
  \item \textbf{The study gives a first detailed account of the drill-down step, where the answer is
    decided.} Most AEO work stops at surfacing; we follow what agents actually do once they read,
    and how well the answer they build holds up.
  \item \textbf{The study leaves an experimental design later work can build on.} AX travels with how
    well known a business already is; we hold fame, prior knowledge, and AEO equal across four
    harnesses, so the AX effect stands on its own.
\end{enumerate}

\section{Related Work}

\paragraph{AEO/GEO: optimizing for surfacing.}
\citet{aggarwal2024geo} showed that page edits (statistics, quotations, citations) can raise a
source's share-of-answer, but \emph{conditional on the source already being retrieved into context}.
Later work tempers this. C-SEO Bench finds most rewrites ineffective and frequently harmful to a
source's ranking, with classic SEO more effective \citep{puerto2025cseo}, and a critical survey of 45
studies concludes that no technique it reviews shows a
stable, longitudinal, cross-platform causal effect on organic discoverability
\citep{geosurvey2026}. We take AEO as governing surfacing only, and show it is inert at the
drill-down.

\paragraph{Agents that browse.}
The pieces were demonstrated separately before the word \emph{agent} settled on them: browsing
and citing sources \citep{nakano2021webgpt}, interleaving reasoning with actions
\citep{yao2023react}, and deciding when to call a tool \citep{schick2023toolformer}. Benchmarks
then established multi-step interaction as standard agent behaviour, on real sites
\citep{deng2023mind2web,he2024webvoyager} and on reproducible self-hosted ones
\citep{zhou2023webarena,mialon2023gaia}. Most on-point,
\citet{gou2025mind2web2} evaluate agentic deep research on correctness \emph{and} source attribution
over live browsing. That is the exact setting we study, with the variable reversed: they compare agents over a
web they let vary, while we hold the agents fixed and compare sites.

\paragraph{Grounding and attribution.}
Retrieval-augmented generation grounds answers in fetched content
\citep{lewis2020rag,guu2020realm,izacard2023atlas}. Attribution (whether an output is
attributable to identified sources) is a formal, evaluable construct \citep{rashkin2023ais},
with benchmarks for answer-with-citations \citep{gao2023alce}, post-hoc attribution
\citep{gao2023rarr}, and faithfulness scoring \citep{es2024ragas}. Our first-party grounding and
answer-composition metrics instantiate these over the specific source the agent read.

\paragraph{Factuality and the cost of ungrounded answers.}
Ungrounded generation hallucinates \citep{ji2023hallucination,huang2023hallucinationsurvey}, and
parametric memory is unreliable for less-popular, long-tail facts, precisely the regime of most
businesses \citep{mallen2023popqa}, and for fast-changing facts \citep{vu2024freshllms}. We score
factual precision atomically against captured ground truth \citep{min2023factscore}. Human audits
find generative search answers frequently unsupported or mis-cited: only \n{51.5\%} of generated
sentences are fully supported by their citations \citep{liu2023verifiability}, and an eight-engine
audit found most responses carried incorrect information \citep{tow2025citation}. Surfacing is not
the same as grounded, accurate answering. These audits grade the engine; we hold the engine fixed
and vary the source it is trying to read.

\paragraph{The risk of grounding on third-party content.}
Content an agent retrieves from pages the business does not control is an attack surface: planted
instructions and text can hijack LLM applications \citep{greshake2023indirect,zou2025poisonedrag},
and steer which products LLM search recommends
\citep{nestaas2024adversarialseo,kumar2024productvisibility,chen2026searchgeo}. Most directly,
\citet{zhang2026ugcpoisoning} poison deep-research agents by editing a single Reddit or Wikipedia
page with as few as 13 words, getting the text cited in 38--51\% of reports. We do not measure
poisoning, but it sharpens our findings: the third-party pages the AEO playbook seeds are the same
channel these attacks exploit, and a readable first-party site is the one source a business can
vouch for.

\paragraph{The agent-readable web.}
Agents arrived on a web that was not built for them, and was closing further. Audits from 2024
found AI crawlers fetching HTML without executing JavaScript, leaving client-rendered content
unreachable
\citep{vercel2024crawler}, while access was being restricted, metered, or blocked by default
\citep{longpre2024consent,cloudflare2025default,cloudflare2025paypercrawl}. What followed has
landed unevenly. The Model Context Protocol caught on and is now the industry standard for
exposing tools and data to agents \citep{anthropic2024mcp}. \texttt{llms.txt}
\citep{howard2024llmstxt} is more widely deployed but lightly used: of agents that read it, only
one in three goes on to fetch a page it lists \citep{agentready2026}. NLWeb
\citep{microsoft2025nlweb} has been published but not taken up. Newer work goes further: WebMCP lets a page expose its
own functions as callable tools \citep{webmcp2025}, and \citet{steiner2026interfaces} find agents
served through structured interfaces (MCP, RAG, and NLWeb) answer more accurately and far more
cheaply than by browsing HTML.
None of these is yet widely adopted, so this study does not explore them. It measures the path
every business already has: an agent fetching and reading the site's own pages. What agent
readiness requires is still being written down, revised as protocols emerge and as more is
measured about how agents behave \citep{agentready2026}.

\section{The Shift: Agents Rely Far Less on Training Data}

Today's agents build their answers from what they fetch, not from what they remember. If a model
still answered mostly from its training data, whether it could read a site would hardly matter.

\begin{figure}[htbp]
  \centering
  \includegraphics[width=\linewidth,%
    alt={Line chart tracking the share of an answer built from training knowledge across OpenAI releases. It falls from 52 percent at gpt-4.1 in 2025 H1, to 37 percent across gpt-5, 5.1 and 5.2, to 27 percent at gpt-5.4 and 5.5, to 14 percent at gpt-5.6, and finally to 7 to 10 percent for this study's four harnesses in 2026 H2. An annotation marks where reasoning models become the default.}]{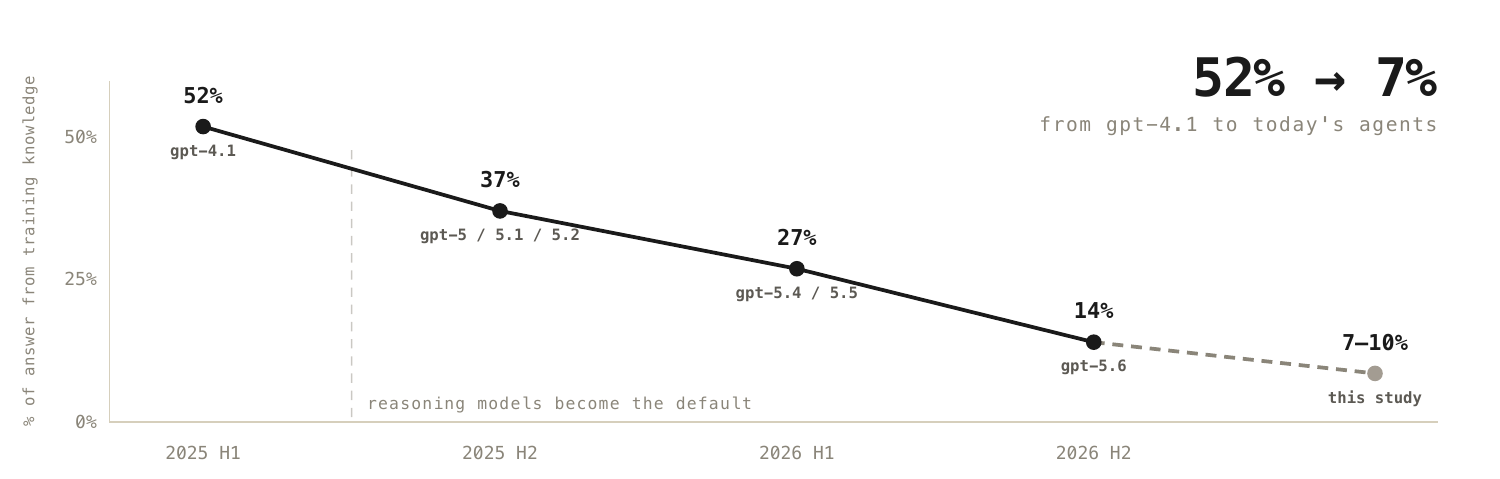}
  \caption{Share of the answer built from training knowledge across seven OpenAI releases: the same
  90 buyer questions, web tools optional, every claim judged. Points are release half-year averages
  over 630 runs. The dashed final point is this study's own measurement across four harnesses, a
  different instrument.}
  \label{fig:shift}
\end{figure}

We put 90 buyer questions, the three intents of Section~\ref{sec:design} asked about 30 of the
study's businesses, 15 per group, as plain questions rather than instructions to check the site, to
each OpenAI generation in turn, web tools available but optional, and had a judge
(\texttt{claude-sonnet-4-6}) score how much of each answer came from memory rather than from retrieval. The last non-reasoning flagship, \texttt{gpt-4.1}, still built about
half its answer from memory. Then reasoning models became the default and it falls off a cliff:
\n{14\%} by \texttt{gpt-5.6}, and \n{7--10\%} in the four agent harnesses this study runs,
measured with a different instrument (Figure~\ref{fig:shift}).

\begin{figure}[htbp]
  \centering
  \includegraphics[width=\linewidth,%
    alt={Three panels of public data. Left: the share of ChatGPT searches anchored to the current year rises from 6 percent on GPT-5.2 to 53 percent on GPT-5.3 and 87 percent on GPT-5.5. Centre: 9 in 10 recency-dependent questions trigger a live web search. Right: Reddit's daily share of ChatGPT citations falls from 3.8 percent on 7 July to 0.5 percent on 17 August.}]{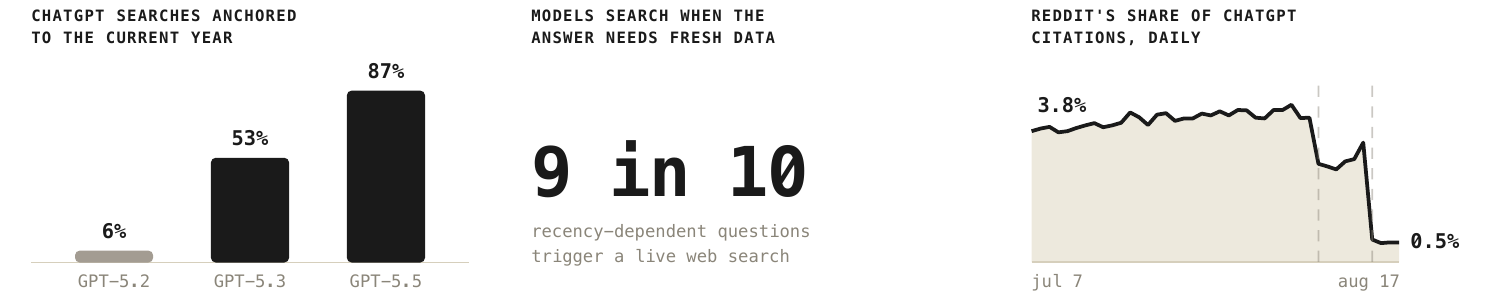}
  \caption{Left: searches anchored to the current year \citep{maestra2026seo}. Centre: search
  invocation on recency-dependent questions \citep{kale2025lookitup}. Right: Reddit's daily share
  of ChatGPT citations \citep{promptwatch2026reddit}.}
  \label{fig:public}
\end{figure}

The public record agrees (Figure~\ref{fig:public}). On recency-dependent questions, frontier models
invoke search about nine times in ten \citep{kale2025lookitup}. Across three ChatGPT generations
between January and May 2026, the share of Business-tier runs that fired a web search whose
searches are anchored to the current year rose from \n{6\%} to \n{87\%} \citep{maestra2026seo}. And the sources those
searches return are themselves a moving target: Reddit's share of ChatGPT citations fell \n{86--95\%}
within a week in August 2026, depending on the panel
\citep{promptwatch2026reddit,qwairy2026reddit}, as \texttt{site:} queries against
official domains went from near zero to roughly a quarter of the model's background searches
\citep{qwairy2026reddit}. An off-site channel can lose its value in a single model update, and
this update moved the model toward the business's own domain.

Memory still answers stable facts in a plain chat turn, and we do not claim otherwise. But buyer
questions about pricing, setup, and comparisons need current information, and in agent harnesses
fetching is the default. In that setting, what the agent reads now decides nearly all of the answer.
The rest of this paper measures what decides what it can read.

\section{Method}
\label{sec:design}

\subsection{Design}

Businesses are drawn from the ora database, one domain per company. The study contrasts two groups differing on a
single axis, agent readiness, matched on the confounds that could otherwise produce the same
result: third-party citations, fame, and the model's prior knowledge of the brand. A difference that
survives matching on all three cannot be explained by them.

Each business is measured on four dimensions, one treatment and three controls held equal across
groups (Table~\ref{tab:balance}).
\begin{itemize}[leftmargin=1.4em,itemsep=3pt,topsep=3pt]
  \item \textbf{Agent readiness (treatment).} The \emph{accessibility} layer of ora's own production
    ranker \citep{ora2026score} in isolation, from its snapshot of 24 August 2026: how fetchable and
    readable a site is to an agent, scored 0--1. A fraction of what the layer checks: how
    much content survives a fetch with no JavaScript; structured data and entity linking;
    machine-readable entry points such as \texttt{llms.txt}, per-URL Markdown fallbacks, HTTP
    \texttt{Link} headers and an agent-discovery file; whether pricing and API documentation are
    reachable; and whether bot controls admit user-triggered agents. Throughout
    the rest of this paper \emph{agent-ready} means accessibility $\geq 0.65$ and \emph{not
    agent-ready} $\leq 0.50$, the study's own cut on this layer rather than the ranker's overall
    grade; the band between is excluded at the candidate stage. The score also serves the
    dose--response analyses.
  \item \textbf{AEO (control).} Two independent proxies. \emph{Citation breadth} scores 0--1 how
    widely a brand is mentioned on sites it does not own: Tavily \citep{tavily2026search}
    queries for \emph{review},
    \emph{alternatives} and \emph{vs} give the distinct third-party hosts, half the score, with a
    quarter each for Wikipedia and Reddit. These are raw search hits, never an answer engine's, so
    the control cannot inherit the outcome it holds fixed. The \emph{discovery score}, another
    layer of the same ranker, measures findability rather than readability. Both are matched
    between groups and tested again as covariates (Section~\ref{sec:results}).
  \item \textbf{Fame (control).} Global traffic rank from the Tranco top-1M list, snapshot of
    24 August 2026 \citep{pochat2019tranco}.
    The cohort excludes the few hundred largest sites: at that size the comparable sites are all
    agent-ready, so there is nothing to pair them against.
  \item \textbf{Prior knowledge (control).} A 0--1 score of how present a brand already is in
    training data, from tool-free probes that map domain to brand and back, recall facts, and name
    the brand within its category.
\end{itemize}

Each agent-ready business is paired with its nearest not-agent-ready neighbour on the standardized
controls, accessibility held out of the distance so matching cannot narrow the contrast. Pairs form
within industry vertical and fame band under a 2.5 standard-deviation caliper, and collection is
blocked by pair. The fame bands split at Tranco rank 110,000, and the 12 verticals are the
categories of Table~\ref{tab:bycat}. Accessibility is re-scanned at launch, so a label reflects
the site on the day it ran.
The result is \n{1,056} businesses, \n{528} per group (Table~\ref{tab:balance}).

\begin{table}[htbp]
\centering
\small
\caption{Covariate balance in the final cohort. The groups differ sharply on the treatment and are
indistinguishable on every control.}
\label{tab:balance}
\sbox\tabbox{\begin{tabular}{llrrr}
\toprule
\textbf{Variable} & \textbf{Role} & \textbf{agent-ready} & \textbf{not agent-ready} & \textbf{SMD} \\
\midrule
Accessibility score      & treatment       & 0.737 & 0.319 & 3.66 \\
\midrule
$\log_{10}$ Tranco rank  & fame            & 5.10  & 5.11  & $-$0.01 \\
Training-data presence   & prior knowledge & 0.332 & 0.332 & 0.00 \\
Citation breadth         & AEO             & 0.322 & 0.327 & $-$0.02 \\
ora discovery score      & AEO             & 0.326 & 0.313 & 0.07 \\
\bottomrule
\end{tabular}}%
\usebox\tabbox
\tabnote{Group means over 528 businesses per group. SMD is the standardized mean difference,
(agent-ready $-$ not agent-ready) / pooled SD; $|\text{SMD}| < 0.1$ is the conventional threshold
for balance.}
\end{table}

\subsection{Collection}

Each business receives three site-anchored prompts, one per intent: pricing, features, and setup.
Each is a natural user request built from a frozen template with the business's own domain filled
in, for example: \emph{``I'm looking for a service that fits my budget. Can you find out what
subscription options are available at \{domain\} and what they cost?''} Every prompt runs three
times on each of the four harnesses, so that run-to-run variance can be separated from differences
between businesses, under a neutral system prompt carrying no behavioural instruction that could
contaminate the experiment.

The four harnesses are independent runtimes spanning two model vendors and two search providers
(Table~\ref{tab:harnesses}), so that no result depends on one agent's implementation. Every run is
spawned in a fresh isolated environment, carrying no state from any other. Each agent can search
the web and fetch pages, using the harness's own tools where it provides them, and must answer.

\begin{table}[htbp]
\centering
\small
\caption{The four agent stacks. Model and search backend vary together, so the table supports no
comparison between one component and another. Tavily is a web-search API for agents
\citep{tavily2026search}.}
\label{tab:harnesses}
\begin{tabular}{lll}
\toprule
\textbf{Harness} & \textbf{Model} & \textbf{Web search} \\
\midrule
claude-agent-sdk & claude-sonnet-4-6 & built-in (Anthropic) \\
claude-code      & claude-haiku-4-5  & built-in (Anthropic) \\
openclaw         & gpt-5.4-mini      & Tavily \\
eve              & gpt-5.4           & Tavily \\
\bottomrule
\end{tabular}
\end{table}

Factual accuracy is measured on a subsample. A full headless-browser render captures what each site
publishes during the collection window, so ground truth exists even for sites the agent was blocked
from reading; the \n{131} businesses whose capture yielded extractable facts for at least one intent
are graded.

\subsection{Measures}

The agent's \emph{trajectory} is the full step-by-step trace of a run. From it and the finished
answer we derive the measures below; judges are blind to a business's group.

\paragraph{Grounding.}
\emph{First-party evidence share} is the fraction of retrieved content (fetched pages plus search
snippets) from the business's own site, in characters. The \emph{grounded-answer rate} is the share
of journeys that read the site and used no outside source. We also record searches, on-site blocks
(HTTP errors or bot walls), turns, duration, and cost.

\paragraph{Answer composition.}
Logs show what the agent read, not what the answer used. A judge splits each answer into sections
and labels each one's source: the business's pages, a third-party page, search snippets, or
training knowledge (only when nothing retrieved could have supplied it).

\paragraph{Independent recommendation.}
A raw recommendation rate would confuse a good product with a readable site, so we hold the business
fixed and ask of each answer how strongly it recommends that business. Two judges from different
vendors (claude-sonnet-4-6 and gpt-5.4) see only the request and the answer and rate it 0 to 4; an
answer \emph{clearly recommends} only when both give the top score. Two vendors guard against a
judge favouring its own family, which also powers harnesses here; they agree within one point
on \n{88\%} of answers. A judge also flags the four \emph{hedge patterns} of
Figure~\ref{fig:hedge}.

\paragraph{Accuracy.}
Each answer is graded against the captured facts one at a time, following the atomic-fact
decomposition of \citet{min2023factscore}: correct $=1$, partial $=\tfrac{1}{2}$ where the answer
gets the fact right but less precisely, incorrect or omitted $=0$. FActScore scores each fact as
supported or not over the facts an answer \emph{states}; we allow half credit and divide by the
facts \emph{asked}, so omission counts as a failure. Graded accuracy is
(correct $+ \tfrac{1}{2}$ partial) / facts asked, over \n{31,127} facts in \n{2,499} answers; an
\emph{empty answer} mentions none. An answer is \emph{site-built} when the agent read the
business's own pages and used no outside source, \emph{web-built} otherwise.

\paragraph{Analysis.}
\label{sec:metrics}
The nine journeys on one business are not independent, so we average within business and treat the
business as the unit of analysis. Group differences are ratios with two-sided permutation tests
(20,000 shuffles) and bootstrap 95\% CIs; trends along the accessibility score use Spearman
correlation. With about 20 comparisons we treat $p < 0.001$ as confirmed and
$0.001 \leq p < 0.05$ as present but not confirmed. AEO proxies are partial correlations
conditioning on accessibility, fame, and the matching strata. Site-built and web-built answers are
compared only within cells of one business, harness, and question type, by sign-flip permutation
(Table~\ref{tab:significance}).

\section{Results}
\label{sec:results}

\FloatBarrier
\subsection{Where the agent actually goes}

\begin{figure}[htbp]
  \centering
  \includegraphics[width=0.90\linewidth,%
    alt={Two side-by-side agent traces for the same buyer question, each split into the business's own site and the open web. Left, twilio.com, graded A: the agent reads five pages on twilio.com and help.twilio.com and answers from Twilio's official docs after a single web search. Right, hashicorp.com, graded D: two hashicorp.com pages are refused by a bot block, the agent falls back to six web searches, and ends on two competitors' blogs at infisical.com and scalr.com.}]{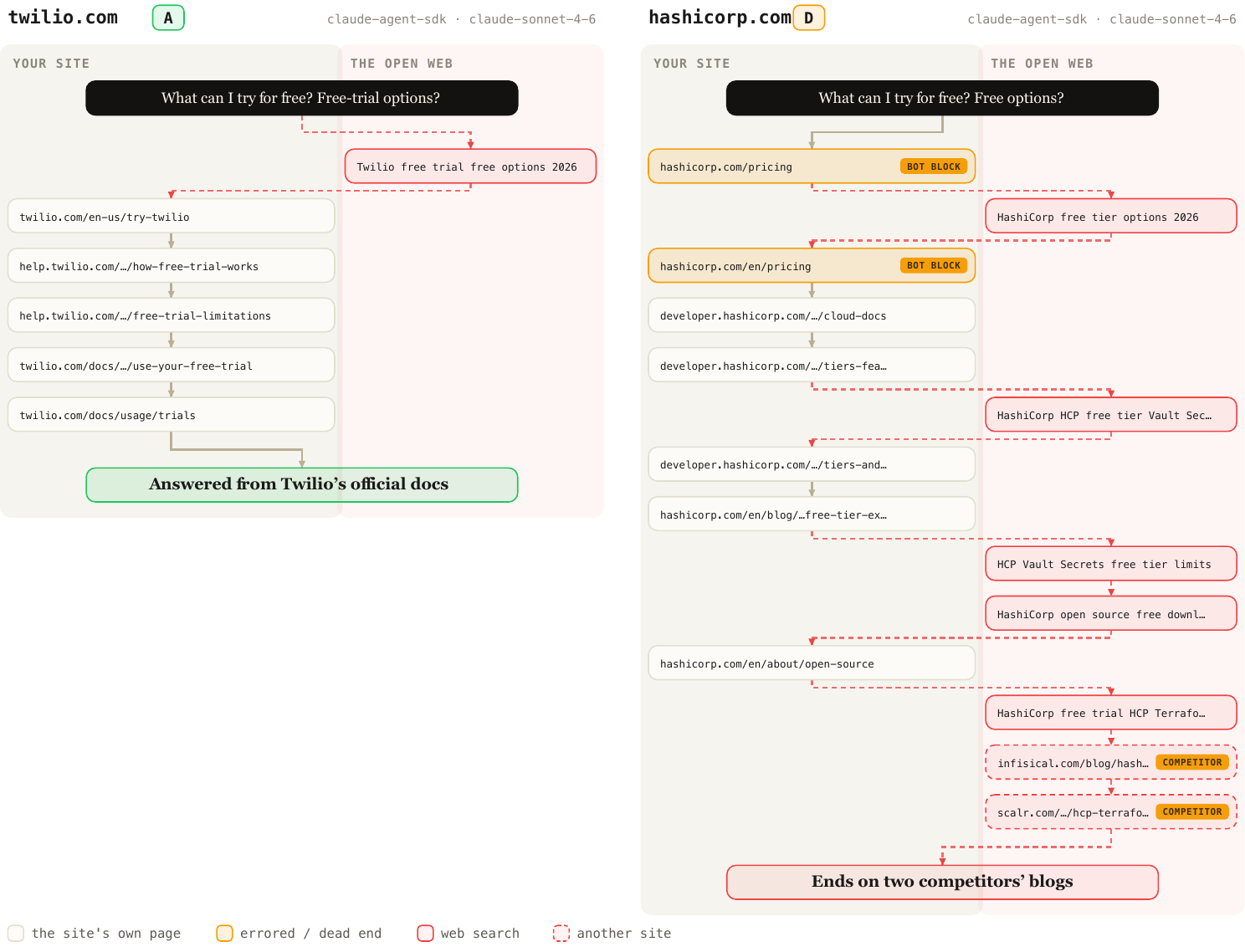}
  \caption{Two real traces, one buyer question, one harness. Left: \texttt{twilio.com},
  agent-ready. Right: \texttt{hashicorp.com}, not agent-ready, and still confident.}
  \label{fig:journeys}
\end{figure}

Figure~\ref{fig:journeys} shows two real traces of the same buyer question on the same harness:
\texttt{twilio.com} answers from its own documentation in seven steps; \texttt{hashicorp.com}
blocks the agent twice, searches the open web seven times, and ends on two competitors' blogs.

\FloatBarrier
\subsection{Recommendation: agent-ready businesses get recommended}

\begin{figure}[htbp]
  \centering
  \includegraphics[width=\linewidth,%
    alt={Two horizontal bars. 20 percent of answers clearly recommend an agent-ready business, against 11 percent when the business is not agent-ready.}]{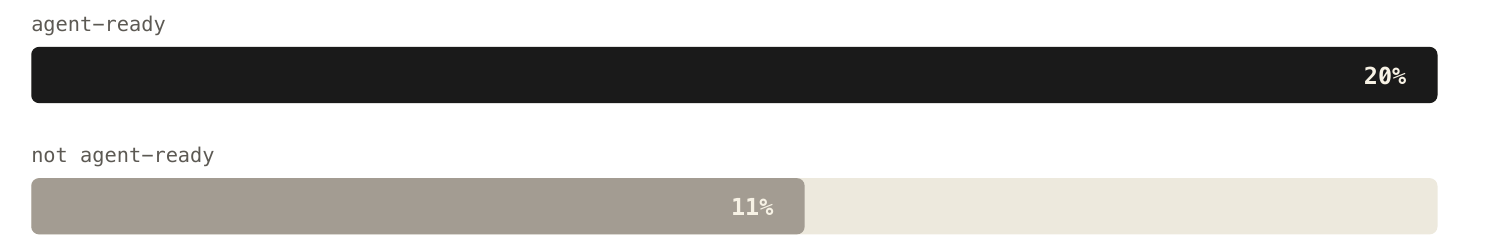}
  \caption{One in five answers clearly recommends an agent-ready business; one in nine when it is
  not agent-ready. Bars are the share of answers that both blind judges rate a clear recommendation,
  all four stacks pooled; both judges must award the top grade for an answer to count (permutation
  $p < 0.0001$).}
  \label{fig:rec}
\end{figure}

A clear recommendation is the outcome the business is competing for: it is what the agent tells the
person who asked. On the two-judge measure, agent-ready businesses get one \n{20\%} of the time
against \n{11\%}, or \n{1.9$\times$} ($p < 0.0001$; Figure~\ref{fig:rec}). Both judges must award
the top grade, so the measure counts only answers that leave the buyer in no doubt.

\begin{figure}[htbp]
  \centering
  \includegraphics[width=\linewidth,%
    alt={Paired bars comparing the clear-recommendation rate for agent-ready and not-agent-ready businesses in each stack. All stacks pooled over 37,927 runs: 20 against 11 percent, a 1.9 times gap. claude-code on claude-haiku-4-5: 5 against 2 percent, 2.6 times. claude-agent-sdk on claude-sonnet-4-6: 14 against 5 percent, 2.6 times. openclaw on gpt-5.4-mini: 25 against 14 percent, 1.8 times. eve on gpt-5.4: 36 against 20 percent, 1.8 times.}]{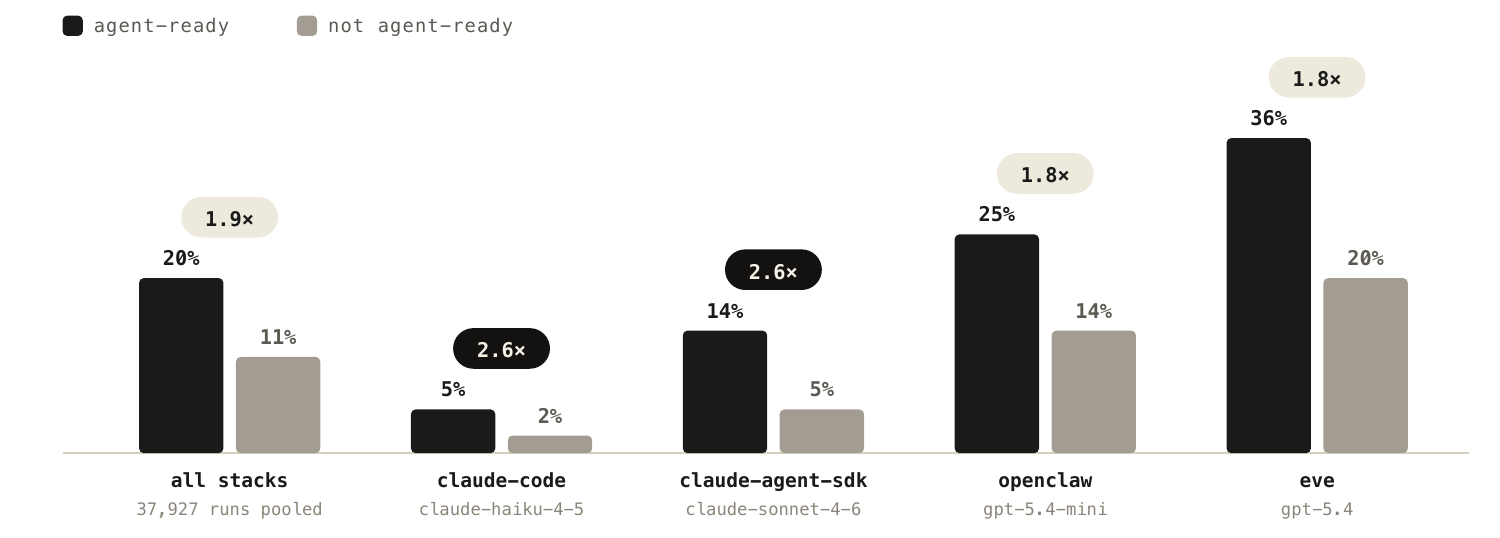}
  \caption{Recommendation rate by agent stack. Bars are the share of answers both blind judges rate
  a clear recommendation.}
  \label{fig:recstack}
\end{figure}

The lift is not one agent's personality. Every stack shows it, but the levels they sit at differ
enormously: claude-code clearly recommends \n{5\%} of the time, eve \n{36\%}
(Figure~\ref{fig:recstack}). That spread belongs to the harness rather than to the businesses, and
it is why the ratio and the absolute movement rank the stacks in opposite orders. claude-code's
\n{2.6$\times$} is three points of movement on a two-point base; eve's \n{1.8$\times$} is sixteen
points. The direction is what replicates across all four.

Every one of the twelve categories shows a lift, from \n{1.2$\times$} in customer service to
\n{2.5$\times$} in IT infrastructure (Table~\ref{tab:bycat}), and it is largest in the categories
where a buyer compares vendors before committing.

\begin{table}[htbp]
\centering
\small
\caption{Recommendation rate by business category, all twelve, ordered by lift.}
\label{tab:bycat}
\setlength{\tabcolsep}{4pt}
\begin{minipage}{0.48\linewidth}
\centering
\begin{tabular}{lrrr}
\toprule
\textbf{category} & \textbf{ready} & \textbf{not} & \textbf{lift} \\
\midrule
IT Infrastructure & 24\% & 9\% & 2.5$\times$ \\
Sales \& Marketing & 21\% & 9\% & 2.3$\times$ \\
Security & 17\% & 7\% & 2.3$\times$ \\
Artificial Intelligence & 23\% & 10\% & 2.2$\times$ \\
Development & 32\% & 14\% & 2.2$\times$ \\
Consumer & 21\% & 9\% & 2.2$\times$ \\
\bottomrule
\end{tabular}
\end{minipage}
\hfill
\begin{minipage}{0.48\linewidth}
\centering
\begin{tabular}{lrrr}
\toprule
\textbf{category} & \textbf{ready} & \textbf{not} & \textbf{lift} \\
\midrule
Analytics & 21\% & 10\% & 2.1$\times$ \\
Fintech \& Payments & 13\% & 8\% & 1.7$\times$ \\
HR \& Legal & 13\% & 8\% & 1.6$\times$ \\
Collaboration \& Prod. & 24\% & 16\% & 1.5$\times$ \\
Commerce & 15\% & 12\% & 1.3$\times$ \\
Customer Service & 23\% & 20\% & 1.2$\times$ \\
\bottomrule
\end{tabular}
\end{minipage}
\widetabnote{Lifts are computed from unrounded rates and will not always reproduce from the
rounded percentages shown.}
\end{table}

\FloatBarrier

\begin{figure}[htbp]
  \centering
  \includegraphics[width=\linewidth,%
    alt={Paired bars for four hedges, agent-ready against not agent-ready. Admits it could not find or access the information: 4 against 16 percent, 4.4 times. Vouches from secondhand sources and cites aggregators instead of the business: 5 against 15 percent, 3.0 times. Vague, noncommittal recommendation with no concrete prices, steps or links: 7 against 12 percent, 1.8 times. Punts the user to go check themselves: 15 against 22 percent, 1.4 times.}]{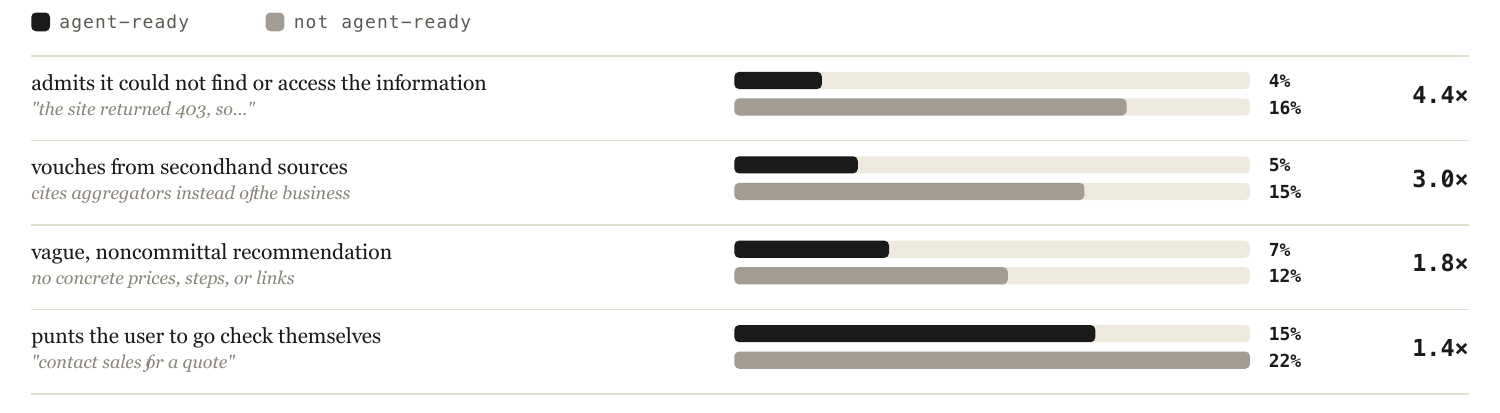}
  \caption{The four ways an answer undermines its own recommendation. Bars are the share of answers
  exhibiting each hedge. All four contrasts are significant at permutation $p < 0.0001$.}
  \label{fig:hedge}
\end{figure}

The same gap appears from the other end. Answers that \emph{both} judges rate too weak to recommend
at all are \n{2.45$\times$} more common when the site is not agent-ready ($p < 0.0001$;
Table~\ref{tab:significance}), and the language of those answers shows why. When an agent cannot
read a site it hedges, and every hedge pattern is significantly more common
(Figure~\ref{fig:hedge}): it discloses that it could not access the business \n{4.4$\times$} more
often, vouches from secondhand sources \n{3.0$\times$} more, is vague \n{1.8$\times$} more, and
sends the user off to check for themselves \n{1.4$\times$} more. An answer that opens by admitting
it could not reach the site is not a recommendation, whatever it says after that.

\FloatBarrier
\subsection{Grounding: when agents can't read you, the answer is built without you}

\begin{figure}[htbp]
  \centering
  \includegraphics[width=\linewidth,%
    alt={Two stacked bars showing what the finished answer is built from. Agent-ready: 78 percent the business's own pages, 3 percent external sites, 12 percent web search, 7 percent training knowledge. Not agent-ready: 58 percent own pages, 7 percent external sites, 25 percent web search, 10 percent training knowledge.}]{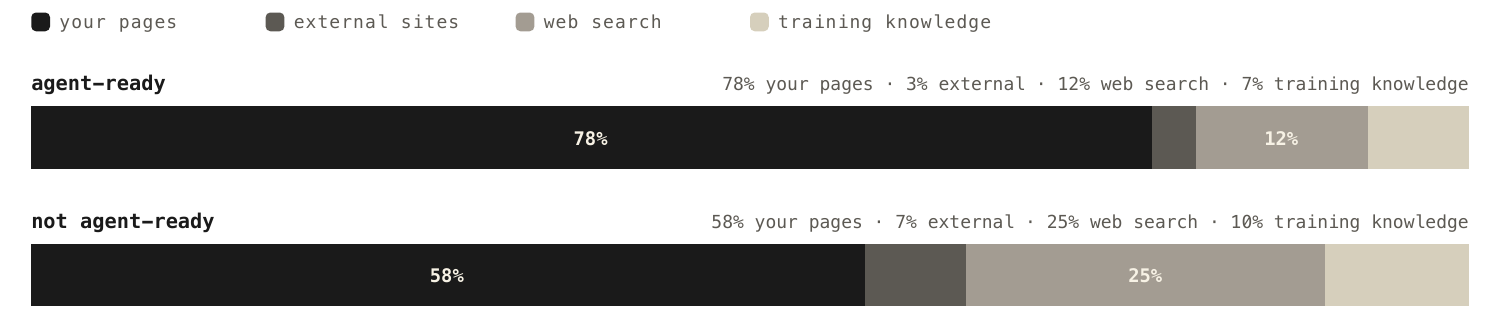}
  \caption{What the finished answer is built from, as shares of the finished text: the business's
  own pages, external sites, web search, and training knowledge.}
  \label{fig:composition}
\end{figure}

In \n{$\sim$99\%} of journeys that hit a dead end on a site, the agent answered anyway, built from
whatever it found elsewhere. Across finished answers, \textbf{training knowledge stays at 7--10\% either way}
(Figure~\ref{fig:composition}), so agents are \emph{not} falling back on memory. Whatever the site
does not provide, web search fills in. For agent-ready businesses, 78\% of the answer is built from
their own pages and only 12\% from web search; when the site is not agent-ready, first-party content
drops to 58\% and web search \textbf{doubles to 25\%}: on sites that are not agent-ready, 42\% of
the answer comes from somewhere other than the business. First-party evidence share is 0.776 vs.\
0.549 (1.41$\times$, $p < 0.0001$).

\begin{figure}[htbp]
  \centering
  \includegraphics[width=0.75\linewidth,%
    alt={Line chart of mean web searches per run against the ora accessibility score, on a zero-based y axis with ticks at 0, 2 and 4. It falls from 4.5 and 4.0 at the least accessible scores, through 3.1, 3.0 and 2.7, to 2.3, 2.1, 2.0 and 1.8 at the most accessible, so the least accessible sites draw 2.6 times more web searches. The middle band from 0.50 to 0.65 is marked excluded.}]{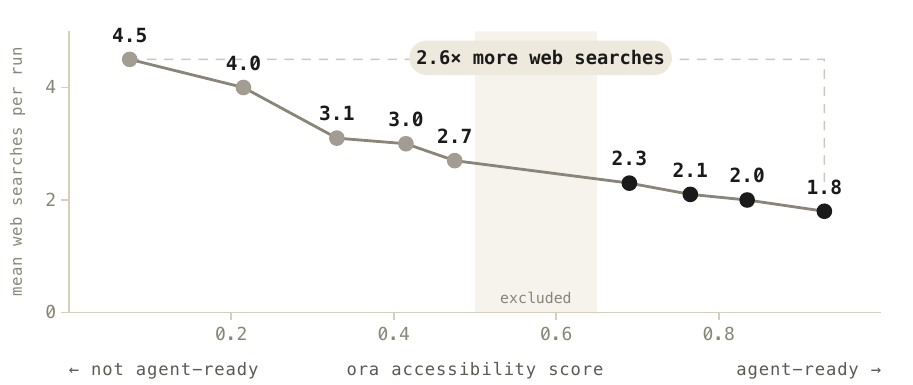}
  \caption{Mean web searches per agent run against the ora accessibility score (Spearman
  $\rho = -0.52$, $p < 0.0001$). The ambiguous middle band of scores, 0.50 to 0.65, is excluded by
  design.}
  \label{fig:searches}
\end{figure}

The mechanism is a clean dose--response in web search (Figure~\ref{fig:searches}): across nine bins
of the accessibility score, mean web searches per journey rises monotonically from 1.8 to 4.5, a
\n{2.6$\times$} spread (Spearman $\rho = -0.52$, $p < 0.0001$, $n = 1{,}056$ businesses). Every
search is another chance the agent is fed information the business never published, or stopped
updating, or that someone planted \citep{zhang2026ugcpoisoning,nestaas2024adversarialseo}.

\FloatBarrier

The pattern survives every cut (Table~\ref{tab:searches}). Search reliance is a stack personality:
openclaw navigates by search and averages \n{6.9} searches even on an agent-ready site, while
claude-code averages \n{0.1}. Losing readability still pushes every stack the same way, by
\n{1.5$\times$} to \n{3.5$\times$}, and every one of the twelve categories moves with it, from
\n{1.2$\times$} to \n{2.0$\times$}.

\FloatBarrier

\begin{table}[htbp]
\centering
\small
\caption{Web searches per run, agent-ready $\to$ not agent-ready: all four stacks, and the five
categories of twelve with the steepest lift.}
\label{tab:searches}
\begin{minipage}{0.48\linewidth}
\centering
\begin{tabular}{lrrr}
\toprule
\textbf{stack} & \textbf{ready} & \textbf{not} & \textbf{lift} \\
\midrule
claude-agent-sdk & 0.4 & 0.9  & 2.4$\times$ \\
claude-code      & 0.1 & 0.3  & 3.5$\times$ \\
openclaw         & 6.9 & 10.4 & 1.5$\times$ \\
eve              & 1.2 & 1.9  & 1.6$\times$ \\
\bottomrule
\end{tabular}
\end{minipage}
\hfill
\begin{minipage}{0.48\linewidth}
\centering
\begin{tabular}{lrrr}
\toprule
\textbf{category} & \textbf{ready} & \textbf{not} & \textbf{lift} \\
\midrule
Consumer                & 1.9 & 3.7 & 2.0$\times$ \\
IT Infrastructure       & 2.0 & 3.9 & 1.9$\times$ \\
Development             & 1.7 & 3.2 & 1.8$\times$ \\
Artificial Intelligence & 1.8 & 3.1 & 1.7$\times$ \\
Sales \& Marketing      & 2.2 & 3.6 & 1.6$\times$ \\
\bottomrule
\end{tabular}
\end{minipage}
\widetabnote{Lifts are computed from unrounded means and will not always reproduce from the
rounded values shown.}
\end{table}

\FloatBarrier
\subsection{Cost: blocked sites cost the agent more per answer}

On sites that are not agent-ready, runs take \n{23\%} more turns (5.5 $\to$ 6.8), get blocked by the
site \n{2.1$\times$} as often, and take \n{15\%} longer to reach an answer (48s $\to$ 55s). All are
significant at $p < 0.0001$ (Table~\ref{tab:significance}). The extra turns and retries are billed to
whoever runs the agent.

\begin{figure}[htbp]
  \centering
  \includegraphics[width=\linewidth,%
    alt={Cost per grounded answer rises from agent-ready to not agent-ready in every stack. openclaw on gpt-5.4-mini: 0.135 to 0.260 dollars, up 93 percent. claude-agent-sdk on claude-sonnet-4-6: 0.112 to 0.216 dollars, up 93 percent. claude-code on claude-haiku-4-5: 0.020 to 0.031 dollars, up 57 percent. eve on gpt-5.4: 0.091 to 0.101 dollars, up 11 percent.}]{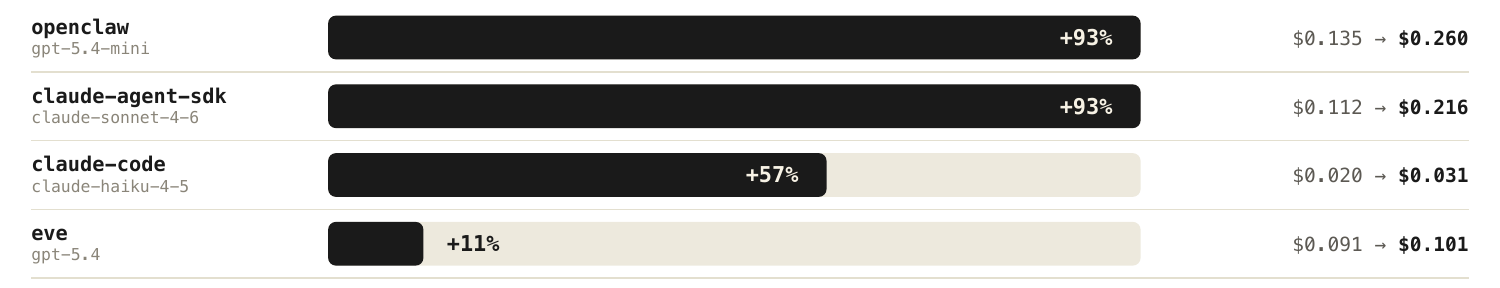}
  \caption{Cost per grounded answer by stack, agent-ready $\to$ not agent-ready.}
  \label{fig:cost}
\end{figure}

Cost per run understates it. A 403 is cheap; the expensive part is that only \n{56\%} of runs on
not-agent-ready sites end grounded in the business, against \n{78\%} on agent-ready sites. Spreading
the cost of the failed runs over the answers that did use the site gives \textbf{+64\% per grounded
answer} averaged across the four stacks, from \n{+11\%} on eve to \textbf{+93\%} on two others
(Figure~\ref{fig:cost}). That premium is paid by whoever operates the agent, not by the business
whose site produced it.

Per run the stacks split, because cost tracks how much the agent reads (Spearman
$\rho = 0.74$--$0.90$ within each stack) and a not-agent-ready site gives back less: loaded pages
carry \n{11--24\%} less text, and runs reading nothing from the site rise from \n{1--12\%} to
\n{11--33\%}. Three stacks search around the gap, read more, and pay \n{6--36\%} more per run. On eve,
the heaviest reader, web search makes up only a quarter of the site text it loses, so it reads
\n{16\%} less in total: its runs get \n{7\%} cheaper but answer from less of the business.

\subsection{Accuracy: the common failure is a missing fact}
\label{sec:accuracy}

The agent picks whether to read the business's own site, and picks it more often on businesses
that are easy for everyone, so a pooled comparison credits the channel with the difficulty of the business: site-built
setup answers come from businesses averaging \n{28.0\%} accuracy against \n{17.4\%} for web-built
ones. Every result below is therefore paired within business, harness, and question type
(Section~\ref{sec:metrics}).

\textbf{Reading the site makes the answer 41\% more accurate.} Site-built answers get \n{48.3\%} of
asked facts right against \n{34.3\%} ($p < 0.0001$; Table~\ref{tab:accuracy}), and an answer built
off-site is \n{3.7$\times$} more likely to contain \emph{none} of the facts the buyer asked for.

\begin{table}[htbp]
\centering
\small
\caption{Site-built vs.\ web-built accuracy over 196 matched cells on 90 businesses; each row
keeps only cells containing both kinds of answer. \emph{diff} is site $-$ web, and for empty
answers the ratio is web/site.}
\label{tab:accuracy}
\setlength{\tabcolsep}{5pt}
\begin{tabular}{lrrrrr}
\toprule
 & \textbf{site} & \textbf{web} & \textbf{diff} & \textbf{lift} & \textbf{$p$} \\
\midrule
\multicolumn{6}{l}{\textbf{Graded accuracy}} \\
All answers                & 48.3\% & 34.3\% & $+$14.0 pp & $+$41\%  & $<$0.0001 \\
\addlinespace[3pt]
\multicolumn{6}{l}{\emph{by question type}} \\
\quad pricing              & 60.4\% & 36.9\% & $+$23.5 pp & $+$64\%  & $<$0.0001 \\
\quad features             & 45.6\% & 38.8\% & $+$6.8 pp  & $+$18\%  & 0.018     \\
\quad setup                & 23.9\% & 23.5\% & $+$0.4 pp  & $+$2\%   & 0.87      \\
\addlinespace[3pt]
\multicolumn{6}{l}{\emph{by agent stack}} \\
\quad claude-code          & 48.3\% & 10.0\% & $+$38.3 pp & $+$382\% & $<$0.0001 \\
\quad claude-agent-sdk     & 48.4\% & 36.1\% & $+$12.3 pp & $+$34\%  & 0.022     \\
\quad openclaw             & 44.0\% & 40.0\% & $+$4.0 pp  & $+$10\%  & 0.051     \\
\quad eve                  & 46.1\% & 48.4\% & $-$2.3 pp  & $-$5\%   & 0.39      \\
\midrule
\multicolumn{6}{l}{\textbf{Empty answers} (containing none of the facts asked)} \\
All answers                & 6.7\%  & 25.0\% & $-$18.3 pp & 3.7$\times$ & $<$0.0001 \\
\bottomrule
\end{tabular}
\end{table}

\begin{figure}[!htbp]
  \centering
  \includegraphics[width=\linewidth,%
    alt={Two stacked bars for the fate of every asked fact. Built from the site: 40 percent stated right, 27 percent half right, 4 percent stated wrong, 29 percent never mentioned. Built from the web: 28 percent stated right, 21 percent half right, 6 percent stated wrong, 45 percent never mentioned.}]{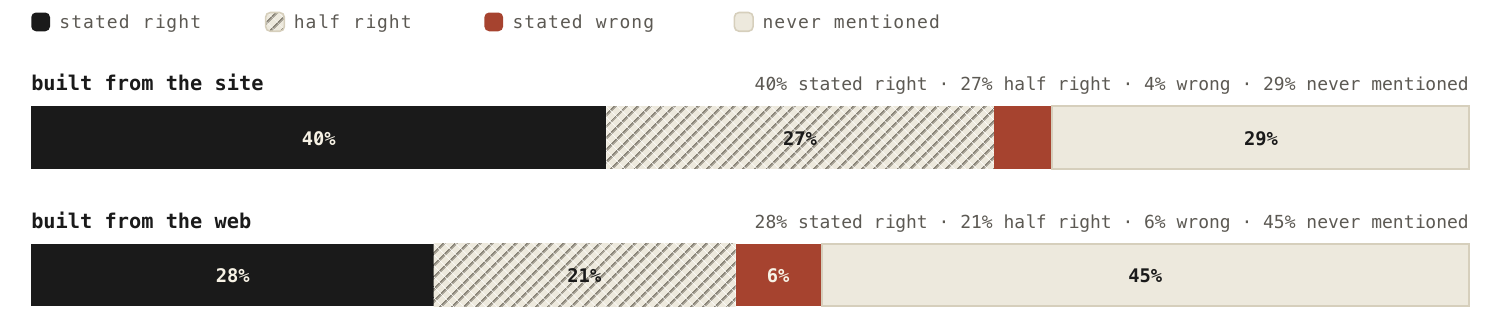}
  \caption{The fate of every asked fact, by where the evidence came from, with shares averaged per
  business so that each business counts once.}
  \label{fig:factfate}
\end{figure}

Fact by fact, the shift is omission rather than error (Figure~\ref{fig:factfate}): stated-wrong
rises only from \n{4\%} to \n{6\%}, while \emph{never mentioned} grows from \n{29\%} to \n{45\%}.
Poor agent readiness makes a fact \textbf{unretrievable}, not false.

The effect is largest on pricing (\n{+64\%}), present on features (\n{+18\%}, $p = 0.018$), and
absent on setup (\n{+2\%}, $p = 0.87$), where \n{65\%} of asked facts go unmentioned in both channels. Setup ground
truth lives in documentation rather than on the marketing pages agents fetch, so what decides the
outcome is findability rather than channel: splitting the setup businesses at their median
site-built share gives \n{31\%} against \n{18\%}, and inside either half the channel barely matters. Across harnesses the gain is largest on the
stack that retrieves least and cannot compensate (claude-code, \n{+382\%}) and disappears on the two
that retrieve most (openclaw, \n{+10\%}, $p = 0.05$; eve, $-$5\%, $p = 0.39$). The ordering follows
how much each stack reads per run: \n{60--75k} characters of evidence for eve and openclaw, \n{4--5k}
for claude-agent-sdk, under \n{2k} for claude-code.

\begin{figure}[htbp]
  \centering
  \includegraphics[width=0.86\linewidth,%
    alt={Graded factual accuracy rises across five bins of the share of the answer's evidence read from the business's own site: 41, 43, 47, 50 and 56 percent from the lowest bin to the highest.}]{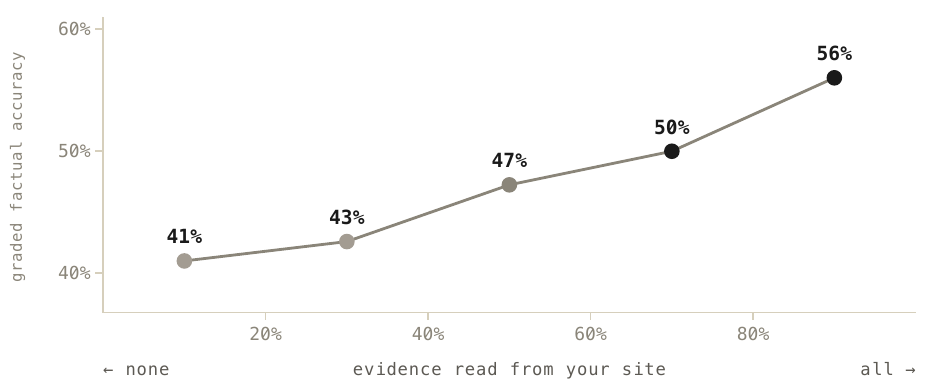}
  \caption{Graded factual accuracy against the share of the answer's evidence read from the
  business's own site, in five bins over graded answers whose run retrieved any content.}
  \label{fig:accuracy}
\end{figure}

Accuracy also tracks grounding as a dose--response (Figure~\ref{fig:accuracy}): grouping answers by
how much of their evidence came from the business's own pages, graded accuracy climbs monotonically
from \n{41\%} to \n{56\%} (Spearman $\rho = 0.19$, $p < 0.0001$).

\paragraph{Group contrast.}
Between assigned groups the graded subset differs by under two points (\n{51.7\%} against
\n{49.8\%}, $p = 0.52$; Table~\ref{tab:significance}). The gain belongs to reading the site, which
readiness makes more likely (\n{82\%} of answers against \n{56\%}) but does not guarantee, so the
group difference is diluted. It is visible: every question type and every harness points the same
way, and on pricing, where the reading effect is largest, it is \n{61.7\%} against
\n{55.6\%}. On 131 businesses none of these reaches significance, and establishing the group
effect directly is left to future work (Section~\ref{sec:limitations}).

\FloatBarrier
\subsection{The AEO control: no effect of its own}

Does AEO matter once readability is known? Across all \n{1,056} businesses, group labels aside,
with accessibility and the other controls held fixed, neither AEO proxy has a confirmed link to
grounding, recommendation, or cost, while accessibility has a strong one to all three
(Table~\ref{tab:aeocontrol}). The one hint, discovery on recommendation ($r = 0.10$,
$p = 0.002$), is below the confirmation bar and fits the thesis: being findable helps a business
get named, not read.

Matching on AEO did not erase the differences this test needs: it equalized the group averages,
but businesses still differ on AEO almost as much as in the full population (\n{93\%} and
\n{92\%} of the spread on the two proxies).

\begin{table}[htbp]
\centering
\small
\caption{The AEO control. Each cell is a partial correlation controlling the accessibility score,
$\log$-Tranco, industry vertical, and fame band; the accessibility column
controls fame and the strata only.}
\label{tab:aeocontrol}
\sbox\tabbox{\begin{tabular}{lrrr}
\toprule
\textbf{outcome} & \textbf{discovery} & \textbf{citation breadth} & \textbf{accessibility} \\
 & \textbf{(AEO)} & \textbf{(AEO)} & \textbf{(AX)} \\
\midrule
Grounding (first-party share) & $+$0.03 (0.32)  & $-$0.04 (0.22) & $\mathbf{+0.62}$ ($<$0.0001) \\
Recommendation (both judges)  & $+$0.10 (0.002) & $-$0.02 (0.45) & $\mathbf{+0.41}$ ($<$0.0001) \\
Cost per journey              & $-$0.04 (0.15)  & $-$0.01 (0.66) & $\mathbf{-0.27}$ ($<$0.0001) \\
\bottomrule
\end{tabular}}%
\usebox\tabbox
\tabnote{Partial $r$, with $p$ in parentheses. Higher accessibility means
more grounding, more recommendation, and lower cost. Dropping the strata changes no verdict.}
\end{table}

\FloatBarrier
\subsection{Summary}

Table~\ref{tab:significance} collects the headline comparisons in one place: thirteen of the
fourteen separate the two groups at $p < 0.0001$.

\begin{table}[htbp]
\centering
\small
\caption{Statistical significance of the headline comparisons (agent-ready vs.\ not agent-ready,
domain-collapsed; ratio in its natural orientation; two-sided permutation $p$; bootstrap 95\% CI on
the ratio).}
\label{tab:significance}
\sbox\tabbox{\begin{tabular}{lrrrlr}
\toprule
\textbf{Comparison} & \textbf{ready} & \textbf{not ready} & \textbf{ratio} & \textbf{95\% CI} & \textbf{perm.\ $p$} \\
\midrule
First-party evidence share            & 0.776 & 0.549 & 1.41$\times$ & [1.36, 1.47] & $<$0.0001 \\
Grounded-answer rate                  & 0.778 & 0.555 & 1.40$\times$ & [1.34, 1.47] & $<$0.0001 \\
Web searches / journey                & 2.16  & 3.38  & 1.57$\times$ & [1.49, 1.65] & $<$0.0001 \\
Turns / journey                       & 5.54  & 6.81  & 1.23$\times$ & [1.20, 1.26] & $<$0.0001 \\
Duration (s)                          & 47.8  & 55.0  & 1.15$\times$ & [1.12, 1.19] & $<$0.0001 \\
On-site block rate                    & 0.158 & 0.335 & 2.12$\times$ & [1.94, 2.33] & $<$0.0001 \\
Cost / journey (USD)                  & 0.068 & 0.079 & 1.16$\times$ & [1.12, 1.20] & $<$0.0001 \\
Recommended (both judges, top score) & 0.204 & 0.106 & 1.93$\times$ & [1.74, 2.15] & $<$0.0001 \\
Too weak to recommend (both judges)  & 0.050 & 0.124 & 2.45$\times$ & [2.11, 2.84] & $<$0.0001 \\
\midrule
Hedge: could-not-access disclaimer    & 0.036 & 0.159 & 4.39$\times$ & [3.68, 5.26] & $<$0.0001 \\
Hedge: vouches from secondhand sources& 0.050 & 0.151 & 3.05$\times$ & [2.65, 3.52] & $<$0.0001 \\
Hedge: vague / noncommittal           & 0.066 & 0.121 & 1.84$\times$ & [1.60, 2.11] & $<$0.0001 \\
Hedge: punts user to the source       & 0.153 & 0.219 & 1.44$\times$ & [1.31, 1.58] & $<$0.0001 \\
\midrule
Graded accuracy, by group (131 businesses) & 0.517 & 0.498 & 1.04$\times$ & [0.93, 1.17] & 0.52 (n.s.) \\
\bottomrule
\end{tabular}}%
\usebox\tabbox
\tabnote{Dose--response (Spearman): web searches vs.\ accessibility
$\rho = -0.52$ ($p < 0.0001$, $n = 1{,}056$); graded accuracy vs.\ first-party evidence share
$\rho = 0.19$ ($p < 0.0001$, $n = 2{,}425$ answers).}
\end{table}

\section{Conclusions}

Six findings. The recommendation, grounding, and search contrasts hold in every harness and every
business category tested; the accuracy and cost effects are averages that vary by harness.

\begin{enumerate}[leftmargin=1.6em,itemsep=4pt,topsep=4pt]
  \item \textbf{Agents answer from what they fetch, not from what they remember.} Across OpenAI
    releases the share of the answer built from memory fell from \n{52\%} in \texttt{gpt-4.1} to
    \n{14\%} in \texttt{gpt-5.6}, and in the four harnesses here it is \n{7--10\%} whether or not
    the site is agent-ready. Optimizing what the model knows about a business now optimizes a
    shrinking slice.
  \item \textbf{Agent readiness decides what the answer is made of.} When the agent can read the
    site, \n{78\%} of the finished answer comes from the business's own pages and \n{12\%} from
    web search; when it cannot, \n{58\%} and \n{25\%}. A blocked agent does not stop: in
    \n{$\sim$99\%} of journeys that hit a dead end it answered anyway, and searches per journey
    rise monotonically as accessibility falls, from \n{1.8} to \n{4.5}.
  \item \textbf{Agent-ready businesses get recommended.} Both blind judges call the answer a clear
    recommendation \n{20\%} of the time for agent-ready businesses against \n{11\%}, or
    \n{1.9$\times$}. Answers about not-agent-ready businesses hedge instead: they are
    \n{4.4$\times$} more likely to admit the site could not be reached and \n{2.45$\times$} more
    likely to be judged too weak to recommend at all.
  \item \textbf{The failure is omission, not fabrication.} Within the same business, harness, and
    question, site-built answers get \n{41\%} more of the asked facts right on average, and web-built ones
    are \n{3.7$\times$} more likely to contain none of them: wrong facts barely move
    (\n{4\%} to \n{6\%}), unmentioned ones climb from \n{29\%} to \n{45\%}.
  \item \textbf{Not being agent-ready costs the agent first, then the business.} On not-agent-ready
    sites runs take \n{23\%} more turns, hit a block \n{2.1$\times$} as often, and take \n{15\%}
    longer; each answer grounded in the business costs \n{64\%} more on average, from \n{11\%} on
    eve to \n{93\%} on two stacks. The agent pays the premium; the business pays in the lost recommendation.
  \item \textbf{Current AEO practices have no effect of their own.} With accessibility held fixed, neither citation
    breadth nor discovery predicts grounding, recommendation, or cost across all \n{1,056}
    businesses, while accessibility predicts all three. Being talked about does not substitute for
    being agent-ready.
\end{enumerate}

AEO's premise is that an answer engine draws on the sources it is pointed to, so a business should
be present wherever those sources are gathered. That premise stops one step short of where the
answer is now decided. Agents build it from what they fetch: the business's own site when it lets
them, whatever else they find when it does not. With fame, prior knowledge, and AEO held equal,
agent readiness moved what the answer was made of, whether the business was recommended, and what
each answer cost, while the AEO proxies had no effect of their own.

The practical reading is a decomposition, not a replacement: answer-engine optimization is SEO for
the surfacing step plus AX for the drill-down step and everything after it. The current playbook
addresses only the first half, and at the drill-down its proxies were inert in every measure.

Being found still matters. Being read is what decides the answer, and it is the one input the
business owns.

\section{Limitations and Future Work}
\label{sec:limitations}

\paragraph{Content depth is not matched.}
The groups are matched on fame, prior knowledge, and AEO, not on how much a site publishes. A
business that blocks agents may also publish less, and thinner content would widen the gap on its
own. The accuracy comparison is immune, since it compares answers about the same business, and on
the graded subset the not-agent-ready sites do publish substantive facts; the grounding and
recommendation contrasts run between businesses and cannot rule this out. An intervention study,
improving a site's readiness and measuring before and after, would settle it.

\paragraph{Accuracy is graded on a subset.}
Ground truth exists for 131 businesses. Within them, answers built from the site are 41\% more
accurate, and the contrast between the two groups points the same way in every question type and
harness but is not significant at this size (Section~\ref{sec:accuracy}). A larger graded set
would test it directly.

\paragraph{Scope.}
The cohort is SaaS and commerce-heavy and English-first, and the intents are business-fact
lookups. Extending it to more verticals, languages, and transactional tasks
comes next.

\section*{Data availability}
Analysis code and a sample of the derived data are released at
\url{https://github.com/ora/research/tree/main/ax-is-the-new-aeo}.

\bibliography{refs}

@inproceedings{aggarwal2024geo,
  title     = {{GEO}: Generative Engine Optimization},
  author       = {Aggarwal, Pranjal and Murahari, Vishvak and Rajpurohit, Tanmay and others},
  booktitle = {Proceedings of the 30th ACM SIGKDD Conference on Knowledge Discovery and Data Mining},
  year      = {2024},
  url       = {https://arxiv.org/abs/2311.09735}
}

@inproceedings{puerto2025cseo,
  title     = {{C-SEO} Bench: Does Conversational {SEO} Work?},
  author       = {Puerto, Haritz and Gubri, Martin and Green, Tommaso and others},
  booktitle = {Advances in Neural Information Processing Systems (Datasets and Benchmarks Track)},
  year      = {2025},
  url       = {https://arxiv.org/abs/2506.11097}
}

@misc{geosurvey2026,
  title        = {Optimizing Visibility in Generative Engines: A Critical Survey of Generative Engine
                  Optimization (2023--2026)},
  author       = {Martinez, Olivier},
  year         = {2026},
  howpublished = {arXiv preprint arXiv:2607.14035},
  url          = {https://arxiv.org/abs/2607.14035}
}

@inproceedings{yao2023react,
  title     = {{ReAct}: Synergizing Reasoning and Acting in Language Models},
  author       = {Yao, Shunyu and Zhao, Jeffrey and Yu, Dian and others},
  booktitle = {International Conference on Learning Representations},
  year      = {2023},
  url       = {https://arxiv.org/abs/2210.03629}
}

@misc{nakano2021webgpt,
  title        = {{WebGPT}: Browser-assisted Question-answering with Human Feedback},
  author       = {Nakano, Reiichiro and Hilton, Jacob and Balaji, Suchir and others},
  year         = {2021},
  howpublished = {arXiv preprint arXiv:2112.09332},
  url          = {https://arxiv.org/abs/2112.09332}
}

@inproceedings{schick2023toolformer,
  title     = {Toolformer: Language Models Can Teach Themselves to Use Tools},
  author       = {Schick, Timo and Dwivedi-Yu, Jane and Dess{\`i}, Roberto and others},
  booktitle = {Advances in Neural Information Processing Systems},
  year      = {2023},
  url       = {https://arxiv.org/abs/2302.04761}
}

@inproceedings{deng2023mind2web,
  title     = {{Mind2Web}: Towards a Generalist Agent for the Web},
  author       = {Deng, Xiang and Gu, Yu and Zheng, Boyuan and others},
  booktitle = {Advances in Neural Information Processing Systems (Datasets and Benchmarks Track)},
  year      = {2023},
  url       = {https://arxiv.org/abs/2306.06070}
}

@inproceedings{gou2025mind2web2,
  title     = {{Mind2Web} 2: Evaluating Agentic Search with Agent-as-a-Judge},
  author       = {Gou, Boyu and Huang, Zanming and Ning, Yuting and others},
  booktitle = {Advances in Neural Information Processing Systems (Datasets and Benchmarks Track)},
  year      = {2025},
  url       = {https://arxiv.org/abs/2506.21506}
}

@inproceedings{he2024webvoyager,
  title     = {{WebVoyager}: Building an End-to-End Web Agent with Large Multimodal Models},
  author       = {He, Hongliang and Yao, Wenlin and Ma, Kaixin and others},
  booktitle = {Annual Meeting of the Association for Computational Linguistics},
  year      = {2024},
  url       = {https://arxiv.org/abs/2401.13919}
}

@inproceedings{zhou2023webarena,
  title        = {{WebArena}: A Realistic Web Environment for Building Autonomous Agents},
  author       = {Zhou, Shuyan and Xu, Frank F. and Zhu, Hao and others},
  booktitle    = {International Conference on Learning Representations},
  year         = {2024},
  url          = {https://arxiv.org/abs/2307.13854}
}

@inproceedings{mialon2023gaia,
  title        = {{GAIA}: A Benchmark for General {AI} Assistants},
  author       = {Mialon, Gr{\'e}goire and Fourrier, Cl{\'e}mentine and Swift, Craig and others},
  booktitle    = {International Conference on Learning Representations},
  year         = {2024},
  url          = {https://arxiv.org/abs/2311.12983}
}

@inproceedings{lewis2020rag,
  title     = {Retrieval-Augmented Generation for Knowledge-Intensive {NLP} Tasks},
  author       = {Lewis, Patrick and Perez, Ethan and Piktus, Aleksandra and others},
  booktitle = {Advances in Neural Information Processing Systems},
  year      = {2020},
  url       = {https://arxiv.org/abs/2005.11401}
}

@inproceedings{guu2020realm,
  title     = {{REALM}: Retrieval-Augmented Language Model Pre-Training},
  author       = {Guu, Kelvin and Lee, Kenton and Tung, Zora and others},
  booktitle = {International Conference on Machine Learning},
  year      = {2020},
  url       = {https://arxiv.org/abs/2002.08909}
}

@article{izacard2023atlas,
  title   = {Atlas: Few-shot Learning with Retrieval Augmented Language Models},
  author       = {Izacard, Gautier and Lewis, Patrick and Lomeli, Maria and others},
  journal = {Journal of Machine Learning Research},
  volume  = {24},
  number  = {251},
  pages   = {1--43},
  year    = {2023},
  url     = {https://arxiv.org/abs/2208.03299}
}

@article{rashkin2023ais,
  title   = {Measuring Attribution in Natural Language Generation Models},
  author       = {Rashkin, Hannah and Nikolaev, Vitaly and Lamm, Matthew and others},
  journal = {Computational Linguistics},
  volume  = {49},
  number  = {4},
  pages   = {777--840},
  year    = {2023},
  url     = {https://arxiv.org/abs/2112.12870}
}

@inproceedings{gao2023alce,
  title     = {Enabling Large Language Models to Generate Text with Citations},
  author       = {Gao, Tianyu and Yen, Howard and Yu, Jiatong and others},
  booktitle = {Conference on Empirical Methods in Natural Language Processing},
  year      = {2023},
  url       = {https://arxiv.org/abs/2305.14627}
}

@inproceedings{gao2023rarr,
  title     = {{RARR}: Researching and Revising What Language Models Say, Using Language Models},
  author       = {Gao, Luyu and Dai, Zhuyun and Pasupat, Panupong and others},
  booktitle = {Annual Meeting of the Association for Computational Linguistics},
  year      = {2023},
  url       = {https://arxiv.org/abs/2210.08726}
}

@inproceedings{es2024ragas,
  title     = {{RAGAS}: Automated Evaluation of Retrieval Augmented Generation},
  author       = {Es, Shahul and James, Jithin and Espinosa-Anke, Luis and others},
  booktitle = {Conference of the European Chapter of the Association for Computational
               Linguistics: System Demonstrations},
  pages     = {150--158},
  year      = {2024},
  url       = {https://arxiv.org/abs/2309.15217}
}

@inproceedings{min2023factscore,
  title     = {{FActScore}: Fine-grained Atomic Evaluation of Factual Precision in Long Form Text Generation},
  author       = {Min, Sewon and Krishna, Kalpesh and Lyu, Xinxi and others},
  booktitle = {Conference on Empirical Methods in Natural Language Processing},
  year      = {2023},
  url       = {https://arxiv.org/abs/2305.14251}
}

@inproceedings{mallen2023popqa,
  title     = {When Not to Trust Language Models: Investigating Effectiveness of Parametric and Non-Parametric Memories},
  author       = {Mallen, Alex and Asai, Akari and Zhong, Victor and others},
  booktitle = {Annual Meeting of the Association for Computational Linguistics},
  year      = {2023},
  url       = {https://arxiv.org/abs/2212.10511}
}

@inproceedings{vu2024freshllms,
  title     = {{FreshLLMs}: Refreshing Large Language Models with Search Engine Augmentation},
  author       = {Vu, Tu and Iyyer, Mohit and Wang, Xuezhi and others},
  booktitle = {Findings of the Association for Computational Linguistics: ACL 2024},
  pages     = {13697--13720},
  year      = {2024},
  url       = {https://arxiv.org/abs/2310.03214}
}

@article{ji2023hallucination,
  title   = {Survey of Hallucination in Natural Language Generation},
  author       = {Ji, Ziwei and Lee, Nayeon and Frieske, Rita and others},
  journal = {ACM Computing Surveys},
  year    = {2023},
  url     = {https://arxiv.org/abs/2202.03629}
}

@article{huang2023hallucinationsurvey,
  title        = {A Survey on Hallucination in Large Language Models: Principles, Taxonomy,
                  Challenges, and Open Questions},
  author       = {Huang, Lei and Yu, Weijiang and Ma, Weitao and others},
  journal      = {ACM Transactions on Information Systems},
  volume       = {43},
  number       = {2},
  year         = {2025},
  doi          = {10.1145/3703155},
  url          = {https://arxiv.org/abs/2311.05232}
}

@inproceedings{liu2023verifiability,
  title     = {Evaluating Verifiability in Generative Search Engines},
  author    = {Liu, Nelson F. and Zhang, Tianyi and Liang, Percy},
  booktitle = {Findings of the Association for Computational Linguistics: EMNLP},
  year      = {2023},
  url       = {https://arxiv.org/abs/2304.09848}
}

@misc{tow2025citation,
  title        = {{AI} Search Has a Citation Problem},
  author       = {Ja{\'z}wi{\'n}ska, Klaudia and Chandrasekar, Aisvarya},
  year         = {2025},
  howpublished = {Tow Center for Digital Journalism, Columbia Journalism Review},
  url          = {https://www.cjr.org/tow_center/we-compared-eight-ai-search-engines-theyre-all-bad-at-citing-news.php}
}

@misc{zhang2026ugcpoisoning,
  title        = {Deep-Research Agents Can Be Poisoned via User-Generated Content},
  author       = {Zhang, Tingwei and Triedman, Harold and Shmatikov, Vitaly},
  year         = {2026},
  howpublished = {arXiv preprint arXiv:2605.24245},
  url          = {https://arxiv.org/abs/2605.24245}
}

@misc{chen2026searchgeo,
  title        = {How Much Can We Trust {LLM} Search Agents? {Measuring} Endorsement Vulnerability to Web Content Manipulation},
  author       = {Chen, Yimeng and Ren, Zhe and Laakom, Firas and others},
  year         = {2026},
  howpublished = {arXiv preprint arXiv:2606.16821},
  url          = {https://arxiv.org/abs/2606.16821}
}

@inproceedings{nestaas2024adversarialseo,
  title        = {Adversarial Search Engine Optimization for Large Language Models},
  author       = {Nestaas, Fredrik and Debenedetti, Edoardo and Tram{\`e}r, Florian},
  booktitle    = {International Conference on Learning Representations},
  year         = {2025},
  pages        = {4857--4888},
  url          = {https://arxiv.org/abs/2406.18382}
}

@misc{kumar2024productvisibility,
  title        = {Manipulating Large Language Models to Increase Product Visibility},
  author       = {Kumar, Aounon and Lakkaraju, Himabindu},
  year         = {2024},
  howpublished = {arXiv preprint arXiv:2404.07981},
  url          = {https://arxiv.org/abs/2404.07981}
}

@inproceedings{greshake2023indirect,
  title     = {Not What You've Signed Up For: Compromising Real-World {LLM}-Integrated Applications with Indirect Prompt Injection},
  author       = {Greshake, Kai and Abdelnabi, Sahar and Mishra, Shailesh and others},
  booktitle = {Proceedings of the 16th ACM Workshop on Artificial Intelligence and Security},
  year      = {2023},
  url       = {https://arxiv.org/abs/2302.12173}
}

@inproceedings{zou2025poisonedrag,
  title     = {{PoisonedRAG}: Knowledge Corruption Attacks to Retrieval-Augmented Generation of Large Language Models},
  author       = {Zou, Wei and Geng, Runpeng and Wang, Binghui and others},
  booktitle = {USENIX Security Symposium},
  year      = {2025},
  url       = {https://arxiv.org/abs/2402.07867}
}

@misc{biilmann2025ax,
  title        = {Introducing {AX}: Why Agent Experience Matters},
  author       = {Biilmann, Mathias},
  year         = {2025},
  howpublished = {biilmann.blog},
  note         = {Published 28 January 2025},
  url          = {https://biilmann.blog/articles/introducing-ax/}
}

@misc{vercel2024crawler,
  title        = {The Rise of the {AI} Crawler},
  author       = {Zecchini, Giacomo and Moore, Alice Alexandra and Ubl, Malte and
                  Siddle, Ryan},
  year         = {2024},
  howpublished = {Vercel engineering report, with MERJ},
  note         = {Published 17 December 2024},
  url          = {https://vercel.com/blog/the-rise-of-the-ai-crawler}
}

@inproceedings{longpre2024consent,
  title     = {Consent in Crisis: The Rapid Decline of the {AI} Data Commons},
  author       = {Longpre, Shayne and Mahari, Robert and Lee, Ariel and others},
  booktitle = {Advances in Neural Information Processing Systems (Datasets and Benchmarks Track)},
  year      = {2024},
  url       = {https://proceedings.neurips.cc/paper_files/paper/2024/hash/c3738949a80306cc48a8ea8ba0560f9d-Abstract-Datasets_and_Benchmarks_Track.html}
}

@misc{cloudflare2026radartraffic,
  title        = {Bot vs.\ Human Traffic},
  author       = {{Cloudflare Radar}},
  year         = {2026},
  howpublished = {Cloudflare Radar, HTML content},
  note         = {Accessed 22 September 2026},
  url          = {https://radar.cloudflare.com/traffic}
}

@misc{cloudflare2025paypercrawl,
  title        = {Introducing Pay Per Crawl: Enabling Content Owners to Charge {AI} Crawlers
                  for Access},
  author       = {Allen, Will},
  howpublished = {Cloudflare blog},
  year         = {2025},
  note         = {Published 1 July 2025},
  url          = {https://blog.cloudflare.com/introducing-pay-per-crawl/}
}

@misc{cloudflare2025default,
  title        = {Content Independence Day: No {AI} Crawl Without Compensation!},
  author       = {Prince, Matthew},
  year         = {2025},
  howpublished = {Cloudflare blog},
  note         = {Published 1 July 2025},
  url          = {https://blog.cloudflare.com/content-independence-day-no-ai-crawl-without-compensation/}
}

@inproceedings{steiner2026interfaces,
  title     = {{MCP} vs {RAG} vs {NLWeb} vs {HTML}: A Comparison of the Effectiveness and
               Efficiency of Different Agent Interfaces to the Web},
  author    = {Steiner, Aaron and Peeters, Ralph and Bizer, Christian},
  booktitle = {Proceedings of the ACM Web Conference},
  year      = {2026},
  url       = {https://arxiv.org/abs/2511.23281}
}

@misc{webmcp2025,
  title        = {{WebMCP}: Exposing Web Application Functionality as Tools for {AI} Agents},
  author       = {{W3C Web Machine Learning Community Group}},
  year         = {2025},
  howpublished = {Draft Community Group Report; not a W3C Standard},
  note         = {First published 13 August 2025},
  url          = {https://github.com/webmachinelearning/webmcp}
}

@misc{howard2024llmstxt,
  title        = {The \texttt{/llms.txt} File},
  author       = {Howard, Jeremy},
  year         = {2024},
  howpublished = {Proposal, llmstxt.org},
  note         = {Proposed 3 September 2024},
  url          = {https://llmstxt.org/}
}

@misc{anthropic2024mcp,
  title        = {Model Context Protocol},
  author       = {{Anthropic}},
  year         = {2024},
  note         = {Announced 25 November 2024},
  url          = {https://modelcontextprotocol.io/}
}

@misc{microsoft2025nlweb,
  title        = {Introducing {NLWeb}: Bringing Conversational Interfaces Directly to the Web},
  author       = {{Microsoft}},
  year         = {2025},
  note         = {Published 19 May 2025},
  url          = {https://news.microsoft.com/source/features/company-news/introducing-nlweb-bringing-conversational-interfaces-directly-to-the-web/}
}

@misc{agentready2026,
  title        = {The Open Standard for Agent Readiness},
  author       = {{agentready.org}},
  year         = {2026},
  howpublished = {Open agent readiness specification, v1.0},
  note         = {Accessed 22 September 2026},
  url          = {https://agentready.org}
}

@misc{kale2025lookitup,
  title        = {Look It Up: Analysing Internal Web Search Capabilities of Modern {LLM}s},
  author       = {Kale, Sahil},
  year         = {2025},
  howpublished = {arXiv preprint arXiv:2511.18931v2},
  note         = {Cited figures are from v2 (28 August 2026)},
  url          = {https://arxiv.org/abs/2511.18931}
}

@misc{maestra2026seo,
  title        = {Is {SEO} Dead in 2026? {Analyzed} 370,000+ Search Results Behind {ChatGPT}
                  and {Gemini}},
  author       = {Kaya, Ali San},
  year         = {2026},
  url          = {https://maestra.ai/blogs/is-seo-dead}
}

@misc{promptwatch2026reddit,
  title        = {Reddit Citations Are Dropping in {ChatGPT}},
  author       = {{Promptwatch}},
  year         = {2026},
  note         = {Published 18 August 2026},
  url          = {https://promptwatch.com/data/reddit-citations-are-dropping-in-chatgpt}
}

@misc{qwairy2026reddit,
  title        = {{ChatGPT} Stopped Citing {Reddit}. {Our} Data Shows the Mechanism Behind
                  the 95\% Collapse},
  author       = {Ilhe, Nicolas},
  year         = {2026},
  note         = {Published 18 August 2026},
  url          = {https://www.qwairy.co/blog/chatgpt-reddit-citations-collapse-august-2026}
}

@inproceedings{pochat2019tranco,
  title     = {Tranco: A Research-Oriented Top Sites Ranking Hardened Against Manipulation},
  author       = {Le Pochat, Victor and Van Goethem, Tom and Tajalizadehkhoob, Samaneh and others},
  booktitle = {Proceedings of the 26th Annual Network and Distributed System Security Symposium
               (NDSS)},
  year      = {2019},
  doi       = {10.14722/ndss.2019.23386},
  note      = {This study uses the Tranco top-1M snapshot of 24 August 2026}
}

@misc{tavily2026search,
  title        = {Tavily: The Web Access Layer for {AI} Agents},
  author       = {{Tavily}},
  year         = {2026},
  url          = {https://tavily.com/}
}

@misc{ora2026score,
  title        = {ora research: agent readiness of the web},
  author       = {{\noopsort{aaa}ora research}},
  year         = {2026},
  note         = {Accessed 27 September 2026},
  url          = {https://ora.ai/research}
}

\end{document}